\documentclass[pmlr]{jmlr}
\jmlrvolume{}
\jmlryear{2019}
\jmlrworkshop{
}

\usepackage[utf8]{inputenc} 
\usepackage[T1]{fontenc}    
\usepackage{amsfonts}       
\usepackage{nicefrac}       
\usepackage{microtype}      

\graphicspath{{images/}}
\usepackage{booktabs}       
\usepackage{multirow}       
\usepackage{multicol}
\usepackage{caption}
\usepackage{subcaption}

\usepackage[titletoc,title,page]{appendix}

\usepackage[shortlabels]{enumitem}
\newlist{todolist}{itemize}{2}
\setlist[todolist]{label=$\square$}

\usepackage[colorinlistoftodos]{todonotes}

\usepackage{adjustbox}
\usepackage{array}
\newcolumntype{R}[2]{%
    >{\adjustbox{angle=#1,lap=\width-(#2)}\bgroup}%
    l%
    <{\egroup}%
}

\title[Feature Robustness in Non-stationary Health Records]{Feature Robustness in Non-stationary Health Records: \\ {\large Caveats to Deployable Model Performance in Common Clinical Machine Learning Tasks}}

\author{%
  \Name{Bret Nestor}\footnotemark[1]
  \Email{bretnestor@cs.toronto.edu}
  \addr \\ \textit{University of Toronto, Vector Institute}
  \AND
  \Name{Matthew B. A. McDermott}\footnotemark[1]
  \Email{mmd@mit.edu}
  \addr \\ \textit{Massachusetts Institute of Technology}
  \AND
  \Name{Willie Boag}
  \Email{wboag@mit.edu}
  \addr \\ \textit{Massachusetts Institute of Technology}
  \AND
  \Name{Gabriela Berner}
  \Email{gberner@college.harvard.edu}
  \addr \\ \textit{Harvard University}
  \AND
  \Name{Tristan Naumann}
  \Email{tristan@microsoft.com}
  \addr \\ \textit{Microsoft Research}
  \AND
  \Name{Michael C. Hughes}
  \Email{mhughes@cs.tufts.edu} 
  \addr \\ \textit{Tufts University}
  \AND
  \Name{Anna Goldenberg}
  \Email{anna.goldenberg@utoronto.ca}
  \addr \\ \textit{Hospital for Sick Children, University of Toronto, Vector Institute}
  \AND
  \Name{Marzyeh Ghassemi} 
  \Email{marzyeh@cs.toronto.edu} 
  \addr \\ \textit{University of Toronto, Vector Institute}
}

\begin{document}
\phantom{\thanks{These authors contributed equally, and should be considered co-first authors.}}
\maketitle

%
\begin{abstract}
When training clinical prediction models from electronic health records (EHRs), a key concern should be a model's ability to sustain performance over time when deployed, even as care practices, database systems, and population demographics evolve. Due to de-identification requirements, however, current experimental practices for public EHR benchmarks (such as the MIMIC-III critical care dataset) are time agnostic, assigning care records to train or test sets without regard for the actual dates of care.
As a result, current benchmarks cannot assess how well models trained on one year generalise to another. 
In this work, we obtain a Limited Data Use Agreement to access year of care for each record in MIMIC and show that all tested state-of-the-art models decay in prediction quality when trained on historical data and tested on future data, particularly in response to a system-wide record-keeping change in 2008 (0.29 drop in AUROC for mortality prediction, 0.10 drop in AUROC for length-of-stay prediction with a random forest classifier). We further develop a simple yet effective mitigation strategy: by aggregating raw features into expert-defined clinical concepts, we see only a 0.06 drop in AUROC for mortality prediction and a 0.03 drop in AUROC for length-of-stay prediction. We demonstrate that this aggregation strategy outperforms other automatic feature preprocessing techniques aimed at increasing robustness to data drift. We release our aggregated representations and code\footnote{Code can be accessed at \url{https://github.com/MLforHealth/MIMIC_Generalisation}} to encourage more deployable clinical prediction models.
\vspace{.5cm}
%
%
%
%
\end{abstract}

%
\section{Introduction}
The wide-spread adoption of electronic health records (EHRs) in modern healthcare systems has enabled the secondary use of these records to develop machine learning models for mortality risk~\citep{harutyunyan2017multitask}, sepsis treatment~\citep{raghu2017continuous}, and many other promising applications \citep{lim2018disease, rajkomar2018scalable}. 
Due to the sensitive nature of patient information, EHR data is typically de-identified in order to reduce risk to patients prior to its use in research. A well-known example of publicly-available, de-identified EHR data is the MIMIC-III database~\citep{johnson2016mimic}, which contains information about intensive care unit (ICU) patients from the Beth Israel Deaconess Medical Center (BIDMC). 

A crucial step of de-identification is obscuring calendar dates related to care. In the MIMIC-III dataset, dates are shifted into the future between the years ``2100 and 2200'' by a consistent random offset for each patient~\citep{johnson2016mimic}. While this preserves privacy, these practices yield a dataset which cannot be analysed in a temporally consistent way---e.g., training a model on historical data, then evaluating on future data. As a result, the wide literature of prior work on MIMIC-III~\citep{harutyunyan2017multitask,purushotham2017benchmark,choi2017gram} all use time-agnostic evaluation protocols, which do not account for a significant source of error that would affect models during true deployment: namely, the evolution of care practices over time and the resultant concept drift~\citep{concept_drift2010}, which are known to induce significant differences in clinical data~\citep{rajkomar2018scalable,lazer2014parable}. These changes can range from mild, gradual drift, such as the typical evolution of care practices and population demographics, to near-instantaneous dramatic shifts such as when the underlying EHR data management system at BIDMC was changed from Philips CareVue\footnote{\url{https://mimic.physionet.org/mimicdata/carevue/}} to MetaVision\footnote{\url{https://mimic.physionet.org/mimicdata/metavision/}} in 2008~\citep{johnson2016mimic}. This shift caused fundamental changes in the way every clinical measurement was recorded in the EHR (yielding entirely new database tables with new variable names).

To the best of our knowledge, researchers have not yet assessed how robust state-of-the-art models trained on MIMIC-III are to temporal drift. In this work, we use a Limited Data Use Agreement allowing restricted access to the underlying calendar year of each event within MIMIC-III to perform such an assessment, examining how well a variety of models generalise to unseen future-only data across a battery of input representations and time-aware training regimes.


We find that the choice of input representation substantially impacts how robust a model is to changing care practices. Models using raw, non-featurised data representations, as advocated by recent deep learning ICU prediction systems such as~\citet{purushotham2017benchmark}, are universally unable to generalise well across large dataset shifts as exemplified by the 2008 system switch within MIMIC-III. Neither dimensionality reduction techniques such as PCA nor automated feature aggregations based on natural language processing are entirely able to circumvent this problem. Across these representations, models trained on only historic data report dramatic drops of area under the receiver-operator curve (AUROC) performance for both a mortality prediction task (worst-case drop of 0.29 AUROC on RF) and a long length-of-stay prediction task (drop of 0.10 AUROC on RF). Such performance problems make the prospective deployment of prediction systems untenable.

To avoid problems when generalising across shifts in feature representations, we introduce a novel \emph{clinically-motivated} feature representation, grouping raw features into underlying concepts, which reduces EHR-shift AUROC drop to 0.06 for mortality, and 0.03 for long length-of-stay. Additionally, by profiling the changes in model performance over time, we find evidence to suggest that both mortality prediction and length-of-stay prediction (each of which are commonly studied tasks) saturate in prediction quality very quickly from little data, suggesting that as a field we should likely focus on more difficult tasks moving forward.

\paragraph{Clinical Relevance}
Clinical data is highly dependent on the landscape of clinical practice as well as underlying population demographics and comorbidities, all of which vary over time. The complete utility of a healthcare model can be nearly impossible to ascertain unless one accounts for the inevitable effect of temporal dataset drift.
However, due to the sensitivity of year-of-care information, this effect has not yet been quantified on MIMIC-III. 
This lack of consideration to the data generating process jeopardises the utility of advancements in clinical machine learning by producing models which are unfit to translate into clinical practice. In this work, we establish how serious this problem is and suggest a mediation of it which should help future models have a better chance of showing robust performance in a real-world clinical setting.

\paragraph{Technical Significance}
In this work, we profile a large number of models across a battery of input representations and several temporal training regimes to assess the robustness of various paradigms to clinical concept drift. We also establish a new, robust representation based on a expert mapping of raw MIMIC-III features into clinical buckets which will be a valuable resource to the machine learning for health community.

%

 

\section{Outline}

\begin{figure}[!t]
    \centering
    \includegraphics[width=\linewidth]{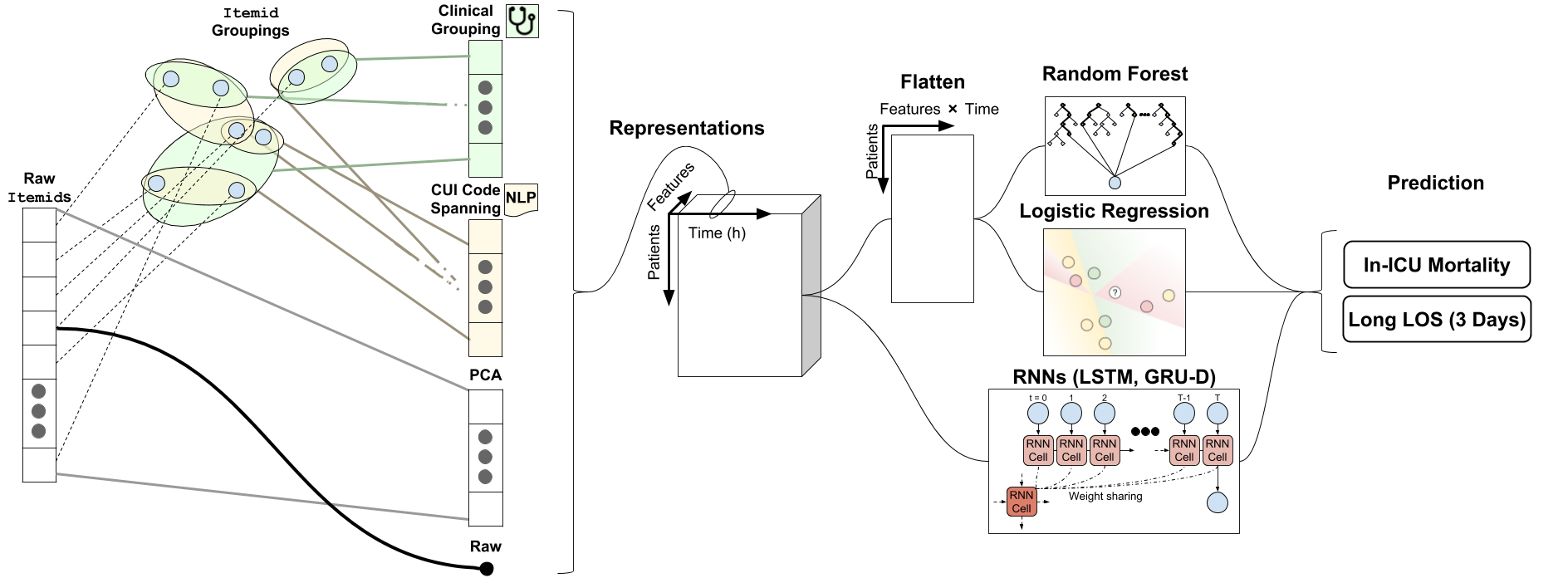}
    \caption{The full experimental pipeline, spanning four data representations, four model types, and two independent prediction tasks. We provide code for reproduction of these results, with the assumption that researchers have obtained the limited-use years data mapping for patient identifiers.}
    \label{fig:full_pipeline}
\end{figure}

We focus on two binary prediction tasks, mortality and long length-of-stay, which are commonly studied for applying machine learning to the MIMIC-III critical care setting. The cohort selection and task setup are described in Section~\ref{sec:data_and_tasks}.
For each task, we evaluate a thorough set of permutations of feature representations, prediction models, and training paradigms (as described in Section~\ref{sec:methods}).
Our full prediction pipeline is illustrated in Figure~\ref{fig:full_pipeline}. We consider four possible representations that span a range from little manual involvement (raw features only) to moderate automatic pre-processing to in-depth expert-selection of high-level features (\textit{Clinical Aggregate}). We further examine four possible models and three possible training regimes that vary how much historical data is included in training.
Results are reported in Section~\ref{sec:results}. Finally, we review related work in generalising across time and clinical sites in Section~\ref{sec:related_work}.



\section{Data Cohort and Prediction Tasks}
\label{sec:data_and_tasks}

\subsection{Data Cohort}
\label{subsec:data}
Within the MIMIC-III dataset, each individual patient may be admitted to the hospital on multiple different occasions, and during each hospital admission may be transferred to and from the intensive care unit (ICU) multiple times. We choose to focus on a patient's first exposure to the ICU (by far the most common case), avoiding the complications of those that transfer multiple times.
We thus extract a targeted cohort of patient EHR data corresponding to the \emph{first} ICU visit. We include only ICU stays that lasted at least 36 hours.
We also focus on non-paediatric case by requiring all patients to be over 15 years old.
These criteria, which broadly follow prior work~\citep{ghassemi2017predicting,suresh2017clinical,mcdermott2018semi},
result in a cohort of $21,877$ unique ICU stays. 

\subsection{Features: Demographics and Hourly Labs and Vitals.} 

For each patient stay, we capture 7 static demographic features and 181 lab and vital measurements that vary over time. The 7 demographic features consist of one-hot encoded gender and race attributes, which are fully observed. The 181 lab results and vital signs have a high-rate of missingness ($>90.6\%$), as each patient may only have a few tests ordered depending on medical needs, and these tests occur infrequently over time.
We used an early version of the MIMIC-III data extraction code made available by \citet{wang2019mimic}.  

\paragraph{Transformation to 24-hour time-series.}
All time-varying measurements from one ICU stay are aggregated into regularly-spaced hourly buckets (0--1 hr, 1--2 hr, etc.). Each recorded hourly value is the mean of any measurements captured in that hour. We normalised each numerical feature to have a mean of zero and a standard deviation of one using a transformation fit on the training dataset. We store this transformation and apply it to the held-out data.

The input to each prediction model contains two parts: the 7 demographic features and an hourly multivariate time-series of labs and vitals, where the feature vector at each hour is determined by the chosen data representation (Section \ref{sec:methods}). We censor the time series to a fixed-duration of 24 hours, representing the first day of a subject's stay in the ICU. 

\paragraph{Imputation of missing values.}
To account for the high rate of missingness within MIMIC-III, we impute our data \cite{janssenMissingCovariateData2010a} via a strategy known as ``simple imputation,'' which was introduced and validated by \citet{cheRecurrentNeuralNetworks2018} for MIMIC time-series prediction tasks. 
Given a chosen representation's observed multivariate time-series, each separate univariate measurement is forward filled, concatenated with a binary indicator if the value was measured within that hour, and concatenated with the time since the last measurement of this value.

\subsection{Binary Prediction Tasks}
We select two representative binary classification tasks for this investigation.
First, the mortality task: given the first 24 hours of data for a patient's ICU stay, predict if the patient will die in the ICU.
Second, the long length of stay (LOS) task: given the first 24 hours of data, predict if the patient will stay in the ICU longer than 3 days.
These tasks are further described in Appendix \ref{app:task_desc}, including previous modelling work and clinical significance.
To prevent label leakage we ensured that all included ICU stays lasted for a minimum duration of 36 hours, which yields a gap of at least 12 hours between when the prediction is made (after the first day) and when the predicted event happens (patient dies or is discharged).

\section{Methods}
\label{sec:methods}
In this section, we outline the various data representations we consider, which prediction methods we assess, and 3 different ways to divide data into training and test sets in chronologically consistent ways to assess how trained models might fare when prospectively deployed. This pipeline is shown in Figure~\ref{fig:full_pipeline}.

\subsection{Representations}
\label{subsec:representation}
Data representation is important for robust model learning~\citep{bengio2013representation}; however, many of the structures that make learning effective in popular ML benchmarks (e.g., Gabor filters for computer vision given natural images) do not map over to clinical equivalents~\citep{raghu2019transfusion}. While others have considered the automated mapping of clinical data elements with mapping tools~\citep{Gong:2017dh} or learned vector space embeddings~\citep{rajkomar2018scalable}, it is unknown whether these methods can withstand time-varying changes in EHR. 

We consider four possible data representations in this work: Raw, PCA, CUI Code Spanning, and Clinical Aggregations. 
These are described in detail below and diagrammed visually in Figure~\ref{fig:full_pipeline}.
Each representation is strictly a way of transforming the feature vector observed at each hour of the observed time-series of labs and vitals (demographics are not affected by the representation).

\paragraph{Raw}
The ``Raw'' representation is the simplest; we include all selected 181 labs and vitals described above in Section~\ref{subsec:data}, each identified via a unique \texttt{ItemID} code in the MIMIC database.
This representation suffers from the significant flaw that the \texttt{ItemID}s are explicitly connected to the underlying EHR software, and thus reflect when logistical practice changes dramatically.
This means any raw features used in the CareVue system before 2008 are \emph{not} used by later MetaVision records, and vice versa. For example, the measurement of ``Heart Rate'' went from \texttt{ItemID} 211 in CareVue to 220045 under MetaVision. In addition to the EHR system shift, there are vitals such as ''Mean Arterial Blood Pressure'' (\texttt{ItemID 6702}) that spontaneously increase their frequency of recording in 2004 (Figure \ref{fig:MABP_freq}). As a result, the ``raw'' hourly time-series are extremely sparse with many missing values before imputation.

%

\paragraph{PCA}
We use principal component analysis~\citep{hotelling1933analysis,jolliffe2002pca} to reduce dimensionality of per-hour raw features.
To train this representation, we provide as input the full \emph{simple imputation} of the 181 raw features, where each of the 181 features has a value, a binary indicator, and a time since last measurement.
Given this 543-dimensional feature vector for each hour of every ICU stay in the training set, we select the first 68 principle components, where 68 was chosen to match the dimensionality of the \textit{clinical aggregate} features. This yields a dense per-hour feature vector with \emph{no} missingness.

\paragraph{CUI Code Spanning}
In a manner similar to that of \citet{Gong:2017dh}, we use the human-readable descriptions of the original $181$ raw \texttt{ItemID}s to automatically aggregate features into groups associated with Concept Unique Identifiers (CUIs) from the Unified Medical Language System (UMLS)~\citep{bodenreider2004unified}. In doing so, a single raw \texttt{ItemID} may be mapped to multiple CUIs, and various raw \texttt{ItemID}s may be mapped to the same CUI; thus, the resulting feature representation reflects an abstraction into a space of shared semantic concepts. This representation yields reduced rates of missingness---though missingness is still present---and has been demonstrated to be more robust across EHR transitions (i.e., when \texttt{ItemID}s change abruptly due to a new EHR)~\citep{Gong:2017dh}. While this representation requires no manual expertise to define the groupings, it relies on the existence of an ontology, UMLS, to provide this CUI to free-text mapping. Note that because multiple CUIs may be identified in the human-readable description of a raw \texttt{ItemID}, we use the \textit{Spanning} pruning technique identified in \citet{Gong:2017dh}, which has been shown to identify the most specific CUIs in a given description~\citep{divita2014sophia}.

\paragraph{Clinical Aggregations}
For this representation, we use expert knowledge to manually define groupings of \texttt{ItemID}s which are converted to a canonical unit space then averaged together. These groupings span the discrepancies between CareVue and MetaVision, such as by grouping 
\texttt{ItemID} values ``Heart Rate'' under CareVue (211) and MetaVision (220045). The groupings also gather together \texttt{ItemID}s which measure the same biophysical quantity merely through different means, such as aggregating MetaVision \texttt{ItemID}s 225664 (``Glucose finger stick''), 220621 (``Glucose (serum)''), and 226537 (``Glucose (whole blood)'') into one unified category for glucose blood sugar. 
The resulting representation groups all 181 raw \texttt{ItemID}s into 68 clinically meaningful categories, and yields a dataset with a rate of 78.25\% missingness before imputation.

\subsection{Models}
\label{sub:models}
To illustrate the effects of non-stationarity, we benchmark four commonly used models in the machine learning literature for clinical prediction from time series data: logistic regression (LR), random forests (RF), long short-term memory (LSTM) network~\citep{hochreiter_long_1997} and a gated recurrent unit with decay (GRU-D) network~\citep{cheRecurrentNeuralNetworks2018}. 
While LSTM and GRU-D can process time series input directly to make a binary prediction for each ICU stay, both LR and RF make predictions using a flattened vector of the 24-hour time-series data.
More details about model implementation can be found in appendix~\ref{app:model_train_details}.

\subsection{Training Regimes}
\label{subsec:training_regimes}
\begin{figure}[h!]
\centering
    \includegraphics[width=0.75\linewidth]{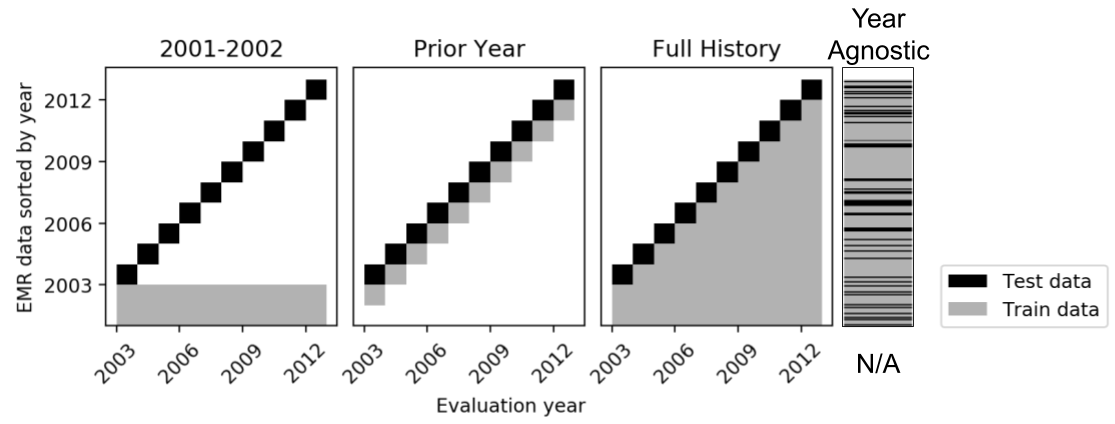}
    \caption{Training paradigms. The available training data for each year is shown in grey. In \textit{2001--2002} data from 2001--2002 is used to build representations and train models that are tested on data from years 2003--2012. In \textit{Prior Year} the data from the year immediately chronologically prior to a given test year is used to train. In \textit{Full History}, all of the available data from 2001 until the year immediately chronologically prior to a given test year is used to train.}
    \label{fig:training_paradigm}
\end{figure}

In addition to measuring the performance of our models when trained in a year-agnostic manner (i.e., the way models are typically run, with no knowledge of the admission year), we use three temporal training paradigms. These training paradigms are designed to capture distinct mechanisms in which practitioners aiming to deploy a clinical model could do so using historical data. We detail these approaches below, and describe them visually in Figure~\ref{fig:training_paradigm}.

\paragraph{Year-Agnostic} Models are trained and tested on randomly shuffled data, with no knowledge of the year of care. This reflects the performance of works reported on publicly available MIMIC data where the true year is not known.
\paragraph{2001--2002} Models are trained on data from 2001--2002 only, and then tested on all future years. This reflects the performance a practitioner can expect when they to train a model on historical data, then deploy it over many years without updating it to reflect more recent data.
\paragraph{Prior Year} For each testing year (from 2003 onward), models are trained on the data from the prior year \emph{only}---e.g., data from 2005 will be used to train a model that is tested on data from 2006. This reflects the performance one can expect if a model were deployed and updated with yearly frequency by re-training from scratch on only the prior year's data. This approach provides the model with the most temporally relevant training data at the expense of training dataset size.
\paragraph{Full History} For each testing year (from 2003 onward), models are trained on \emph{all} prior data---e.g., data from 2001--2005 will be used to train a model that is tested on data from 2006. This reflects the performance one can expect if they deployed a model, updating it yearly by retraining it on all data ever observed. This provides the most available training data, at the expense of temporal significance.

\section{Results}
\label{sec:results}
 
We present the average AUROC per model and representation across all years for the task of early mortality classification in Figure~\ref{fig:mort}. In Table~\ref{tab:MORT_results}, we compare the average ($\pm$ standard deviation) AUROCs and maximum drop in AUROC observed when comparing the first year of evaluation to each subsequent unseen year from 2003 onward using the \textit{Full History} training regime between the \textit{Raw} and \textit{Clinical Aggregation} representations. We summarise the main points of our results below for the task of mortality prediction, and include results for length-of-stay (LOS) in Appendix~\ref{app:los}, Figure~\ref{fig:los}, and Table~\ref{tab:LOS_results}

\subsection{Clinical Aggregate Representations are Most Robust to Non-stationarity}
Models trained on the \textit{Raw} representation suffer a rapid performance decrease in 2008, after the EHR change. We also note that most models do not recover quickly, even those with high capacity (e.g., LSTMs and GRU-D); this is perhaps unexpected under the \textit{Full History} training paradigm. The exception is that the GRU-D begins to recover after receiving approximately 2 years of data. This may suggest GRU-D's explicit handling of missingness offers benefits in response to concept drift. The clinically aggregated representation maintains much more consistent performance across years for all of the models regardless of model complexity. This is demonstrated quantitatively in Table~\ref{tab:MORT_results} as well. It is important to note that clinical aggregation confers much lower standard deviation across all models, which in addition to performance offers more clinically trustworthy models.

PCA performed consistently lower than the \textit{Raw} representation, and struggled to generalise throughout time. Note that we also tested non-linear dimensional reduction with UMAP~\citep{UMAP}, but found similar results: minor perturbations in data distributions from year-to-year renders the model obsolete at test time. Representations that attempt to automatically detect similarities in data elements for grouping managed to sustain both the GRU-D and the logistic regression model to some extent, however failed to help the random forest and LSTM models to recover when trained on the full history of available data.

Overall, the \textit{Clinical Aggregate} representation is most robust to the performance deterioration which is observed for the other representations, surpassing prior work by \citet{Gong:2017dh} with learned representations. 
This suggests that there is room for future research towards automatically generating mappings between multiple EHR systems. 
We also analyse these trends broken out by protected subgroup information, to determine if different subgroups are  susceptible to data drift. We find that, in general, smaller subgroups have (not unexpectedly) higher variance and worse generalisation (see Appendix~\ref{sec:subpop} for more information).

\subsection{Date-Agnostic Training Overstates Performance, Especially in Raw Data}
We replicate the year-agnostic training and test practice common to most reporting in machine learning papers, and found that this method creates an unrealistic upper bound to model performance, \emph{especially} on the raw representation. For example, RF models report a year-agnostic mortality AUROC of $0.82 \pm 0.02$ (5 x 2 fold CV splits~\citep{dietterich1998approximate}), as compared to their true year-averaged AUROC under the raw representation of $0.76 \pm 0.13$. Under the \textit{Clinical Aggregate} representation, in contrast, RF reports a year-agnostic mortality AUROC of $0.86 \pm 0.02$, in comparison to the true year-averaged AUROC of $0.85 \pm 0.02$. A full comparison of static models can be found in Appendix \ref{app:year_agnostic}. The overestimate in performance in the clinical representation is thus only $0.017$, as compared to $0.054$ under the raw representation.

This problem is also especially challenging as increasingly high capacity models are developed and evaluated in a date-agnostic fashion. Prior work in longitudinal EHR data on wound healing noted that the most complex models achieved significantly higher AUROC than all other models only under conditions approximating stationarity~\citep{jung2015implications}. In a non-stationary setting, model complexity had no advantage, and the gap between the best model and the simplest model was substantially reduced. We see similar impacts in this work where lower capacity models (RF, LR) are preferred on both our tasks. 

\subsection{Performance Saturation and Task Importance} 

\begin{figure}
\centering
    \includegraphics[width=1.0\linewidth]{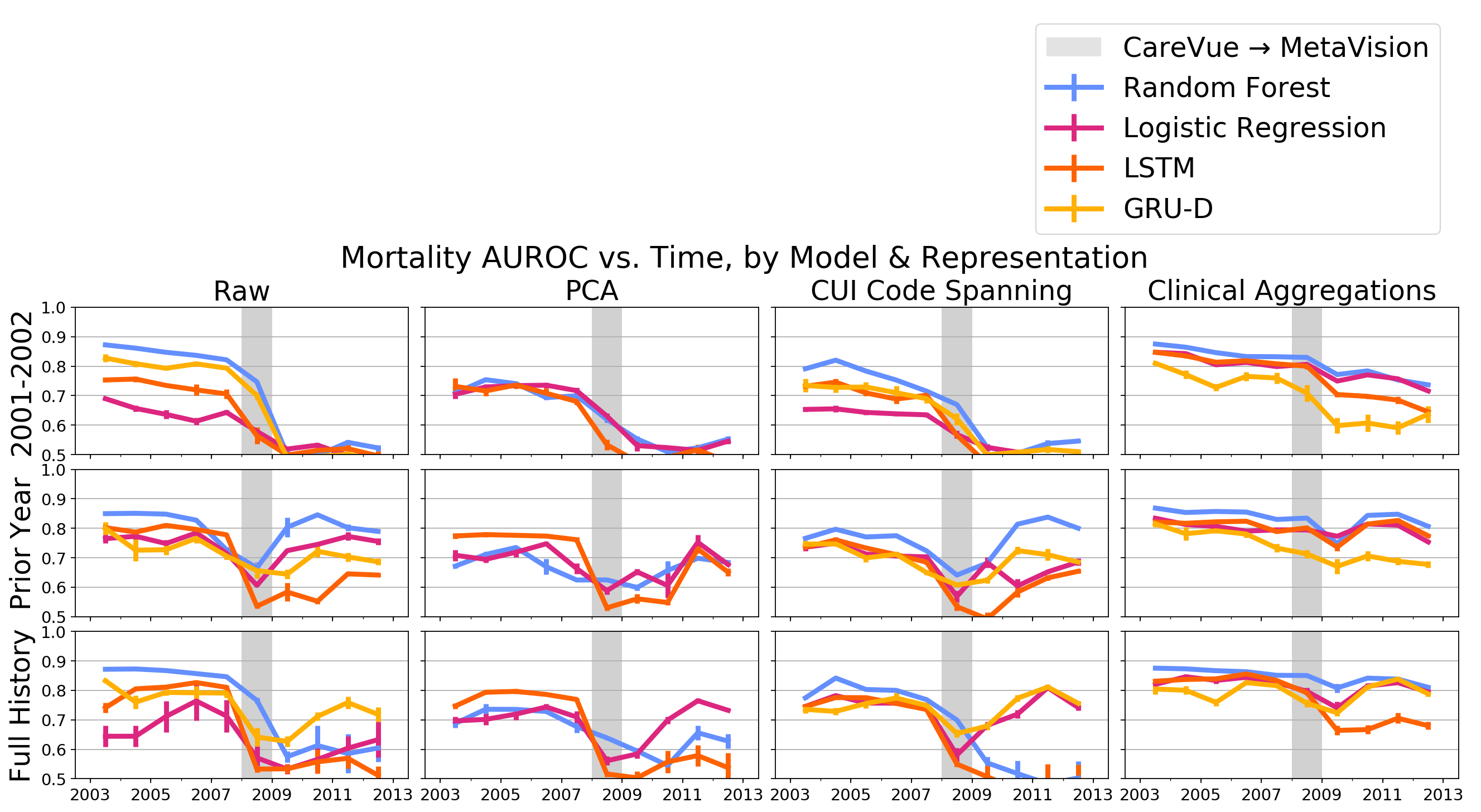}
    \caption{Impact of representation (columns) and training style (rows) on the chronological performance of classifiers for the task of early mortality classification. Error bars indicate $\pm$ standard error. Grey shaded vertical selection indicates transition from CareVue to MetaVision. Note that our \textit{Clinical Aggregate} representation trained on the \textit{full history} is the least deviant and highest performing representation has been attained for most of the models. In contrast, models trained with the \textit{raw} representation tend to have high performance variance before the policy shift  and rapidly deteriorate during the EHR change from CareVue to MetaVision even when trained on the \textit{full history} of data.}
    \label{fig:mort}
\end{figure}

By profiling the changes in model performance over time, we find evidence to suggest that both of the tasks considered (each of which are commonly studied) require relatively few years of aggregated data to saturate in prediction quality. In particular, in Figure~\ref{fig:mort}, under the \textit{Full History} training regime and the \textit{Clinical Aggregates} representation, model performance is very steady from the beginning of the training period where only one year of data is used.
Even as the training dataset grows significantly (increasing tenfold in size from test year 2003 to 2012), performance on unseen future data does not dramatically change. This suggests that as a field we should likely focus on more difficult tasks moving forward, or find alternative signals not already in the measured 181 labs and vitals to improve performance. 


\begin{table}
    \centering
    \caption{
    Mortality prediction task: A comparison of the a) average ($\pm$ standard deviation) AUROC over each unseen year from 2003 onward, and b) max loss observed between the first year of evaluation and subsequent years' performance from 2003 onward. All numbers were computed between the clinical and raw representations under the \textit{Full History} training regime across various models. \textbf{Bold} indicates best performance across all models and representations. Bigger is better for averages, while smaller is better for maximum loss and standard deviation. This table shows that the clinical representation tends to improve the overall performance and decreases the magnitude of performance deterioration during non-stationary healthcare practice.
    }
    
    \begin{tabular}{l || cc || cc} \toprule  
        \multirow{2}{*}{Model} & \multicolumn{2}{c||}{Average AUROC} & \multicolumn{2}{c}{Max AUROC Drop} \\
                               & Raw                             & Clinical                       & Raw   & Clinical                  \\ \midrule
        LR & $0.64 \pm 0.07$ & $\boldsymbol{0.81 \pm 0.03}$  & $0.13$ & $\boldsymbol{0.01}$ \\
        RF & $0.76 \pm 0.13$ & $\boldsymbol{0.85 \pm 0.02}$  & $0.29$ & $\boldsymbol{0.06}$ \\
        LSTM & $0.68 \pm 0.13$ & $\boldsymbol{0.77 \pm 0.07}$  & $0.22$ & $\boldsymbol{0.13}$ \\
        GRUD & $0.75 \pm 0.06$ & $\boldsymbol{0.79 \pm 0.03}$  & $0.18$ & $\boldsymbol{0.07}$ \\
        \bottomrule
    \end{tabular}

    \label{tab:MORT_results}
\end{table}

\section{Related Work}
\label{sec:related_work}
\paragraph{Machine Learning for Health on MIMIC}
Much of the prior work in machine learning for healthcare focuses on the MIMIC-III dataset, which contains electronic health record (EHR) data from $38,600$ adults spanning $58,900$ hospital admissions at Beth Israel Deaconess Medical Center from 2001 to 2012~\citep{johnson2016mimic}.
Researchers have predicted mortality, billing codes, length-of-stay prediction, intervention onset and offset prediction, among other targets \citep{ghassemi2018opportunities,suresh2017clinical,raghu2017continuous,choi2017gram,ghassemi2017predicting,cheRecurrentNeuralNetworks2018}. \citeauthor{harutyunyan2017multitask} and \citeauthor{purushotham2017benchmark} have also defined specific cohorts and benchmarking tasks for MIMIC-III, primarily aiming to enable comparisons of models of varying capacity across meaningful classification tasks. In all cases, authors report aggregate performance on date-agnostic splits into train and test.

\paragraph{Robust Machine Learning over Time}
Works have explored concept drift and proposed specialised methods to account for the resulting data distribution changes in machine learning systems, but predominantly in isolated contexts and typically far from the traditional machine learning for health literature. \citeauthor{davis_calibration_2017}, for example, studies performance and calibration drift in machine learning models trained to predict hospital acquired acute kidney over nine years of care practice evolution. Interestingly, they found that models largely maintained performance, though calibration suffered over a number of model types \citep{davis_calibration_2017}. \citeauthor{beyene_improved_2015} examines concept drift in the task of predicting surgeries, and proposes a novel algorithm to detect significant clinical practice changes and adjust accordingly \citep{beyene_improved_2015}. \citet{subbaswamy2018counterfactual} address dataset shift by estimating causal features and creating a latent space on which to build classification models. 
Another approach estimates a relationship between source and target domains that incorporates prior knowledge of how the data generating process might differ (or stay the same) between domains~\citep{subbaswamy2019preventing}.
The prior two methods do not handle the mapping between multiple EHR systems, hence can be seen as complimentary to the work presented here. Work by \citet{jung2015implications} examines predictive models of wound healing in outpatient care centers over multiple years of data. \citeauthor{jung2015implications} show that gains from non-linear models that capture feature interactions are visible in stationary settings (what we call year-agnostic), but \emph{disappear} when models are evaluated in non-stationary prospective fashion due to covariate shift.
Outside the healthcare domain, \citet{becker2007real} evaluate the use of weighted majority techniques for dealing with concept drift; while their approach considers applications to ranking, the underlying use of ensembles could be adapted to the classification tasks considered here.

Overall, the issue of distributional shift is still considered to be a major barrier in the safety and robustness of ML/AI systems, especially in healthcare \citep{challen_artificial_2019}.

\paragraph{Multi-site Generalisability}
Though examinations of model robustness over \textit{time} are rare in machine learning for healthcare, a number of prior works have focused on the notion of generalisability across healthcare \textit{institutions}. \citeauthor{Gong:2017dh}, for example, uses the clinical Text Analysis Knowledge Extraction System (cTAKES)~\citep{savova2010mayo} to identify UMLS concept unique identifiers (CUIs)~\citep{bodenreider2004unified} from human-readable feature descriptions, and aggregates features into higher level buckets based on this CUI overlap. The resulting representation improves model transfer between the CareVue portion of MIMIC-III (pre-2008) and the Metavision portion of MIMIC-III (post-2008) \citep{Gong:2017dh}. \footnote{This split can be re-identified from the differences in the data directly, though further granularity over time is not generally accessible.}  They frame the CareVue-to-MetaVision transition as a proxy for multiple institution data, not for temporal drift.

Using the recently released multi-source eICU dataset \citep{pollard_eicu_2018}, other work has additionally investigated the use of multi-source training data to create models with robust transferability to novel institutions \citep{johnson_generalizability_2018}. Similarly, \citeauthor{rajkomar2018scalable} trains models using a site-agnostic representation which shows strong predictive performance when transferring between two distinct institutions \citep{rajkomar2018scalable}. Both \citet{johnson_generalizability_2018} and \citet{rajkomar2018scalable} assume \emph{strong} overlap in feature representations, because the underlying EHR systems are the same or there are a sufficiently large number of samples to capture correlations even in the presence of underlying measure sparsity. In this work, we emphasise evaluating how prediction systems are vulnerable to abrupt changes in the EHR system when little overlap exists between representations.
%
%



\section{Conclusion}
By augmenting the popular de-identified MIMIC dataset with non-public year-of-care information obtained via a Limited Data Use Agreement, our experiments have quantified the robustness of state-of-the-art machine learning pipelines to real-world non-stationarity over two common critical care tasks: mortality prediction and length-of-stay prediction. 
We have established that ignoring time during evaluation,
as is common to MIMIC and other public datasets, will consistently overestimate prediction quality and should be regarded sceptically when assessing future deployment potential. Further, we exposed key problems with the robustness of raw feature representations regardless of model type, even in time-aware evaluation scenarios. 
Finally, we have developed novel aggregate representations that are demonstrably more effective at generalising to future years than several automated preprocessing methods. The maximum AUROC performance loss throughout the 2002-2012 in MIMIC-III is reduced from 0.29  to just 0.06 for ICU mortality prediction and from 0.10 to 0.03 for length of stay prediction.

%
This work is intended to serve as a first step towards identifying and overcoming obstacles to effective model deployment. As such, it has several limitations. First and foremost, our present study would not have been possible without the dated events acquired via a Limited Use agreement which others may not have access to immediately. 
Second, our suggested aggregate representations required manual definition specific to the MIMIC database, so transfer to other EHR systems would require new taxonomies to be developed with input from clinical experts. While several existing automated approaches did help, we expect further research could alleviate more of the burden of manual concept definition.  
Finally, we only consider two of the most common tasks used in the MIMIC-III dataset, and more work is needed to understand the impact of non-stationarity in other tasks that may not have such rapid performance saturation. 

We conclude by cautioning researchers in machine learning for healthcare to consider the deployability limits of their models. To benchmark progress in clinical machine learning it is essential to use standardised datasets and tasks, however, we should not expect models trained on date-agnostic, de-identified data to translate into clinical practice.
\label{headings}

\section*{Acknowledgements}
Dr. Marzyeh Ghassemi is funded in part by Microsoft Research, a CIFAR AI Chair at the Vector Institute, a Canada Research Council Chair, and an NSERC Discovery Grant.

Matthew McDermott is funded in part by National Institutes of Health: National Institutes of Mental Health grant P50-MH106933 as well as a Mitacs Globalink Research Award.

\bibliography{references,brett_zotero}

\newpage
\appendix

\section{Model Training Details}
\label{app:model_train_details}
Models which do not implicitly handle missingness (LR, RF and LSTM) require data to be imputed. \citet{cheRecurrentNeuralNetworks2018} detail a thorough list of imputation schemes in their experiments where the \textit{simple imputation} scheme tends to perform well. Alternatively, Gaussian processes have been used to impute missing values in a clinical setting successfully\citep{lasko2013computational}.
Other works have shown that parameterising features by sequence time produces stationary and invariant sets of features\citep{hripcsak2015parameterizing}. To isolate the effect of representation on generalisability we do not explicitly test the effects of imputation schemes on the model performance in this work.

All models have hyperparameters selected via random search \citep{bergstra2012random}, with parameters detailed below. Additional implementation details can be found in the code base: \url{https://github.com/MLforHealth/MIMIC_Generalisation}.

\paragraph{Logistic Regression (LR)} LR models are linear classification models of low capacity and moderate interpretability. 
Because LR does not naturally handle temporal data, 24 one-hour buckets of patient history are concatenated into one vector along with the static demographic vector. 
We use the LR implementation in SciKit Learn's \texttt{LogisticRegression} class~\citep{pedregosa_scikit-learn:_2011}. A model was selected from a random search of regularisation strength (C), regularisation type (L1 or L2), solvers (``liblinear'' or ``saga''), and maximum number of iterations.
\paragraph{Random Forest (RF)} RF models are nonlinear classification models defined using bagged decision trees which are often competitive non-neural, non-linear baseline methods. 
The data is prepared in the same way as the logistic regression model and the RF implementation in SciKit Learn's \texttt{RandomForestClassifier} class is used. A model was selected from a random search of minimum number of samples to split a node, the minimum number of samples per leaf node, maximum depth of the tree, number of estimators in the ensemble.
\paragraph{Long Short-Term Memory (LSTM)} LSTM models \citep{hochreiter_long_1997} are a popular variant of recurrent neural networks (RNNs) capable of processing arbitrary length sequences in a non-linear, high-capacity fashion. We implement a bidirectional LSTM using TensorFlow~\citep{martin_abadi_tensorflow:_2015}. We used a bidirectional LSTM model selected via a random search of dropout (0.1 0.2, 0.3, 0.4, 0.5), number of epochs (1 to 5), hidden layer size (16, 32, 64, or 128 units), activation function (tanh, ReLU), and optimizer (rmsprop, adam, adagrad).
\paragraph{Gated Recurrent Unit with Decay (GRU-D)} GRU-D models \citep{cheRecurrentNeuralNetworks2018} are a recent variant of recurrent neural networks (RNNs) designed to specifically model irregularly sampled timeseries by inducing learned exponential regressions to the mean for unobserved values. Note that as GRU-D is intentionally designed to account for irregularly sampled timeseries (or equivalently timeseries with missingness), evaluating it on representations that internally absorb missingness does not make sense. As such, we do not evaluate GRU-D on the PCA representation. We implemented the model in PyTorch \citep{paszke_automatic_2017} based on \citep{han-jd_inspired_2019,cui_uw_gated_2019}. We use a hidden layer size of 67 units, batch normalisation\citep{ioffe2015batch}, and dropout with a probability of 0.5 on the classification layer like in the original work \citep{cheRecurrentNeuralNetworks2018}. The Adam optimizer \citep{kingma2014adam} is applied with the early stopping criteria\citep{cheRecurrentNeuralNetworks2018}.

\paragraph{Tuning Procedure}
For the RF, LR and LSTM classifiers, 5-fold cross validation was applied to the training data, using a random search to find best parameters for maximum area under the receiver-operator curve (AUROC) on the validation split. For the GRU-D model, we use the early stopping criteria on the validation during training to find the best model for the specified hyper-parameters (variance shown is only attributable to the train/validation split).

\section{Feature aggregation description}
\label{app:feat_agg}
    In MIMIC-III, \texttt{CHARTEVENTS} with \texttt{ItemID}s \textit{861, 1127, 1542, and 220546} are averaged into one feature in the \texttt{Clinically Aggregated} feature vector called \textit{White blood cell count}. Note that \texttt{ItemID}s \textit{861, 1127, and 1542} were recorded in the 2001--2008 system and \texttt{ItfemID} \textit{220546} was recorded in the 2008--2012 system.
    
    A full description of the clinical aggregations can be found in \citet{wang2019mimic}.
    

    \begin{figure}[h]
    \centering
    \includegraphics[width=0.4\linewidth]{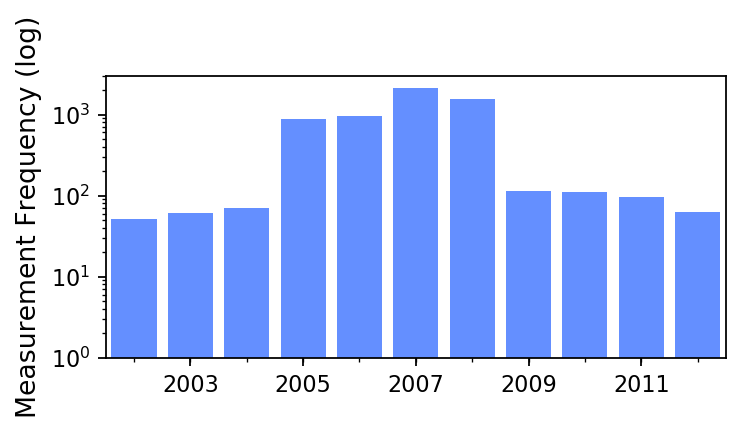}
         \caption{The frequency of data collection can change in clinical practice. Shown is an example of the collection frequency for Mean Arterial Blood Pressure.}
         \label{fig:MABP_freq}
    \end{figure}
    
    \begin{figure}[h]
         \centering 
         \includegraphics[width=0.4\linewidth]{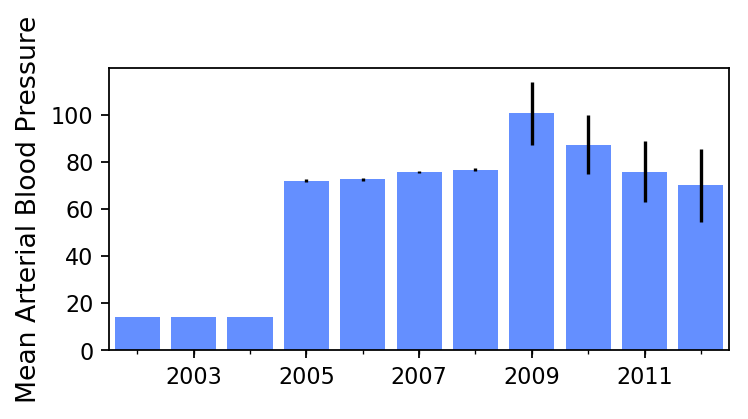}
         \label{fig:y equals x2}
     \caption{The measured values of data can shift in clinical practice}
    \label{fig:example_drift}
    \end{figure}
    
\section{Task Descriptions}
\label{app:task_desc}

\paragraph{In-ICU Mortality Task.} In-ICU Mortality is defined by patient death within the ICU. Mortality prediction in general is a common prediction target \citep{harutyunyan2017multitask,purushotham2017benchmark} as it is a direct signal of acuity strongly associated with EHR signals. The ICU mortality rate of patients in our subset is 7.4\%. We also know that policy changes from the Affordable Care Act led to a changes in clinical practice regarding mortality prevention~\citep{kocher2010affordable}. 

\paragraph{Long LOS Task.} Predicting long LOS has significant utility in clinical operations management, and has been predicted repeatedly in prior work \citep{harutyunyan2017multitask,purushotham2017benchmark}. In this work, we define a ``long LOS'' to be a LOS greater than the median LOS we observe in our cohort (3 days). We use the same data as described previously to predict if a patient will need to be in the ICU for greater than or less than 3 days, e.g., a binary label of 0 if length-of-stay is less than 3 days, otherwise 1. This creates a 47.1\% positive subject rate in-task. This task has the added benefit of investigating date randomisation effects in a more balanced-class problem without a directly targeted policy change. 

\section{Year Agnostic Results}
\label{app:year_agnostic}

Tables \ref{tab:MORT-Year-Agnostic} \& \ref{tab:LOS-Year-Agnostic} contain the model performances when trained without knowledge of the years (5 x 2 fold CV splits~\citep{dietterich1998approximate}).

\begin{table}[h]
    \centering
    \caption{The in ICU Mortality performance trained in a year-agnostic fashion. This represents how machine learning models trained on electronic health records are typically reported in literature. The AUROC (mean $\pm$ std) is reported}
    
    \begin{tabular}{l || cccc c} \toprule  
        Model & \multicolumn{4}{c}{Average AUROC for Random Splits} \\
                              & Raw   & PCA & CUI Code Spanning&  Clinical   \\                    \midrule
        
        LR & $0.71 \pm 0.02$ & $0.79 \pm 0.01$ & $0.68 \pm 0.01$ & $0.85 \pm 0.01$ \\
        RF & $0.82 \pm 0.02$ & $0.77 \pm 0.03$ & $0.79 \pm 0.02$ & $0.86 \pm 0.02$ \\
        LSTM & $0.70 \pm 0.03$ & $0.75 \pm 0.01$ & $0.68 \pm 0.03$ & $0.84 \pm 0.01$ \\
        GRUD & $0.81 \pm 0.04$ & - & $0.80 \pm 0.01$ & $0.83 \pm 0.02$ \\
        \bottomrule
    \end{tabular}
    
    \label{tab:MORT-Year-Agnostic}
\end{table}

\begin{table}[h]
    \centering
    \caption{The in length of stay (greater than 3 days) classification performance trained in a year-agnostic fashion. This represents how machine learning models trained on electronic health records are typically reported in literature. The AUROC (mean $\pm$ std) is reported}
    
    \begin{tabular}{l || cccc c} \toprule  
        Model & \multicolumn{4}{c}{Average AUROC} \\
                              & Raw   & PCA & CUI Code Spanning&  Clinical   \\                    \midrule

        LR & $0.67 \pm 0.02$ & $0.68 \pm 0.01$ & $0.68 \pm 0.01$ & $0.70 \pm 0.01$ \\
        RF & $0.70 \pm 0.00$ & $0.68 \pm 0.01$ & $0.67 \pm 0.01$ & $0.71 \pm 0.01$ \\
        LSTM & $0.65 \pm 0.01$ & $0.62 \pm 0.02$ & $0.63 \pm 0.02$ & $0.69 \pm 0.01$ \\
        GRUD & $0.69 \pm 0.01$ & $ - $ & $0.67 \pm 0.01$ & $0.70 \pm 0.00$ \\
        \bottomrule
        
    \end{tabular}

    \label{tab:LOS-Year-Agnostic}
\end{table}

\section{Performance on Length-of-Stay Task}
\label{app:los}
\begin{figure}[h]
\centering
    \includegraphics[width=1.0\linewidth]{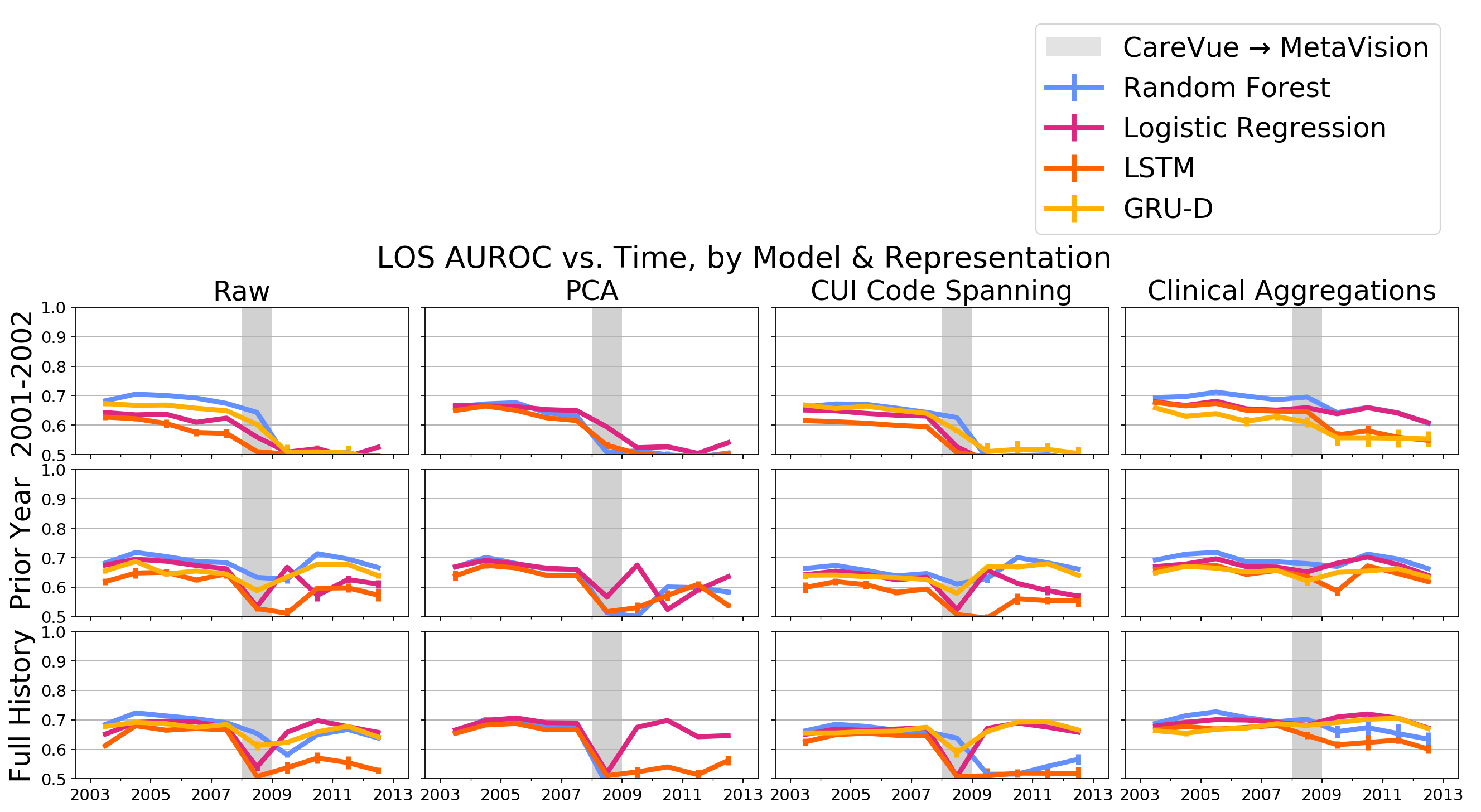}
    \caption{Impact of representation (columns) and training style (rows) on the chronological performance of classifiers for the task of early length of stay classification. Error bars indicate $\pm$ standard error.}
    \label{fig:los}
\end{figure}

We show the full results for the LOS task reported on in the main text of this work here. As noted previously, much of the commentary for mortality holds here as well. Figure~\ref{fig:los} shows the AUROC of various models over time across all our representations and training regimes. Similar to the results in Figure~\ref{fig:mort}, we see dramatic decays in response to the CareVue to MetaVision shift, which are recovered by the clinical aggregation representation. Again, similar to Figure~\ref{fig:mort}, under training regimes that offer insufficient prior data, we see general decay year-over-year as well. 
Table~\ref{tab:LOS_results} compares the average ($\pm$ standard deviation) AUROCs and maximum drop in AUROC observed when comparing the first year of evaluation to each subsequent unseen year from 2003 onward using the \textit{Full History} training regime between the \textit{Raw} and \textit{Clinical Aggregation} representations.

\begin{table}[h]
    \centering
    \caption{
    LOS: A comparison of the a) average ($\pm$ standard deviation) AUROC over each unseen year from 2003 onward, and b) max loss observed between the first year of evaluation and subsequent years' performance from 2003 onward. All numbers were computed between the clinical and raw representations under the \textit{Full History} training regime across various models. \textbf{Bold} indicates best performance across all models and representations. Bigger is better for averages, while smaller is better for maximum AUROC drop. ''$*$'' indicates that 2003 was the worst performing year, and that models consistently improved in subsequent years.
    }
    
    \begin{tabular}{l rr c||c rr} \toprule  
        \multirow{2}{*}{Model} & \multicolumn{2}{c}{Average AUROC} &             & \multicolumn{2}{c}{Max AUROC Drop} \\
                        & Raw                           & Clinical &                      & Raw   & Clinical                      \\ \midrule
        LR & $0.66 \pm 0.04$ & \textbf{0.69 $\pm$ 0.02} & & $0.11$ & $\boldsymbol{* (+0.02)}$ \\
        RF & $0.67 \pm 0.04$ & \textbf{0.68 $\pm$ 0.03} & & $0.10$ & $\boldsymbol{0.03}$ \\
        LSTM & $0.60 \pm 0.06$ & \textbf{0.65 $\pm$ 0.03} & & $0.07$ & $\boldsymbol{0.02}$ \\
        GRUD & $0.66 \pm 0.03$ & \textbf{0.67 $\pm$ 0.03} & & $0.01$ & $\boldsymbol{* (+0.06)}$ \\
        \bottomrule
    \end{tabular}

    \label{tab:LOS_results}
\end{table}

    


\clearpage
\section{Subpopulation Sensitivity Analyses}
\label{sec:subpop}

\begin{figure}[!htb]
\centering
    \includegraphics[width=1.0\linewidth]{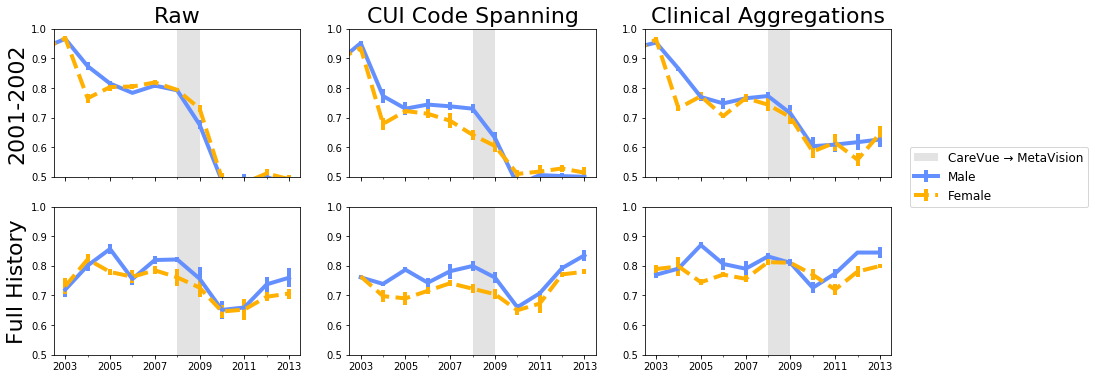}
    \caption{Impact of representation on the longevity of performance across two genders. The model shown is a GRU-D classifying in-ICU mortality.  Error bars indicate $\pm$ standard error.}
    \label{fig:sensA}
\end{figure}

\begin{figure}[!htb]
\centering
    \includegraphics[width=1.0\linewidth]{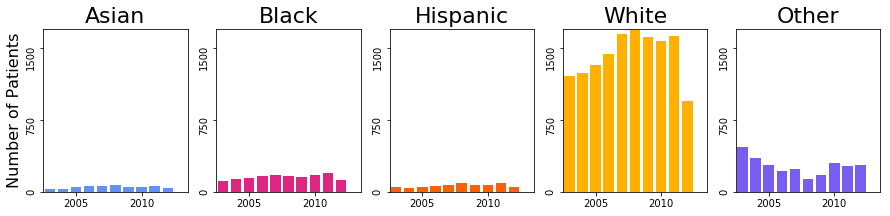}
    \caption{The number of ICU admissions per year by ethnicity.}
    \label{fig:sensC}
\end{figure}

\begin{figure}[!htb]
\centering
    \includegraphics[width=1.0\linewidth]{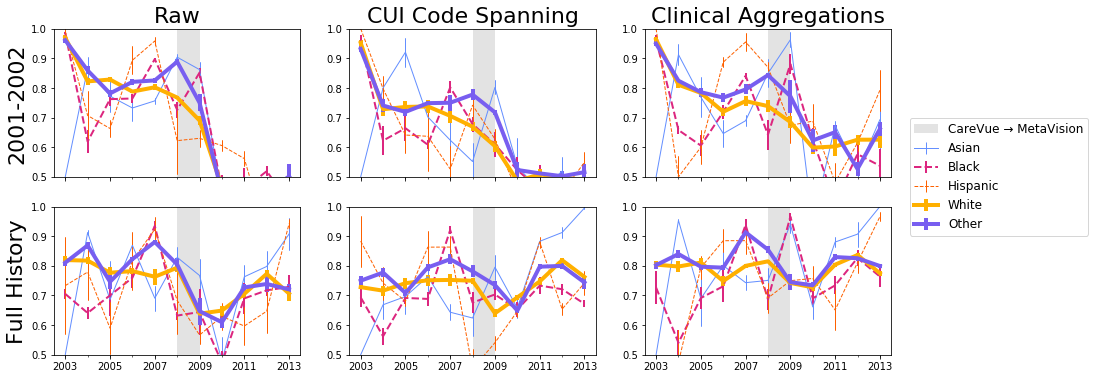}
    \caption{The performance of the GRU-D on the task of mortality prediction. The classification is shown for highlighted demographics.}
    \label{fig:sensB}
\end{figure}

\begin{figure}[!htb]
\centering
    \includegraphics[width=1.0\linewidth]{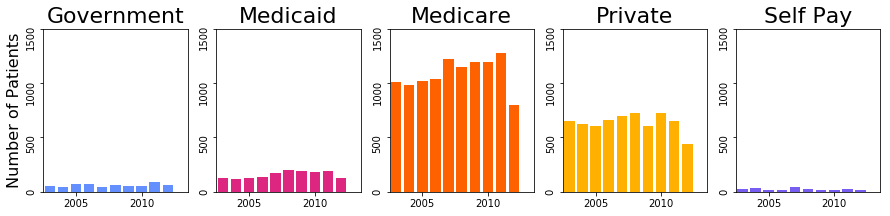}
    \caption{The number of ICU admissions per year by insurance type.}
    \label{fig:sensE}
\end{figure}

\begin{figure}[!htb]
\centering
    \includegraphics[width=1.0\linewidth]{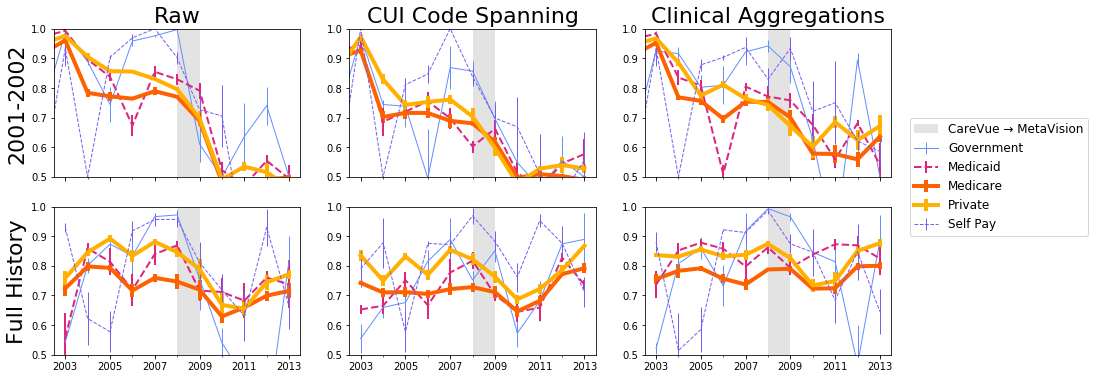}
    \caption{The performance of the GRU-D on the task of mortality prediction. The classification is shown for highlighted insurance types.}
    \label{fig:sensD}
\end{figure}

Here, we consider subpopulation sensitivity. 
In Figure~\ref{fig:sensA} we partition our prediction task across gender.
In Figure~\ref{fig:sensB} we partition our prediction task across ethnicity. A bidirectional LSTM shown on all representations.
In Figure~\ref{fig:sensC} we show the occurrence of each of the sensitive groups from Figure~\ref{fig:sensB}. The predictive performance is erratic for groups with fewer samples.
In Figure~\ref{fig:sensD} we partition our prediction task across insurance type. The performance is demonstrated for the GRU-D model
In Figure~\ref{fig:sensE} we show the occurrence of each insurance type from Figure~\ref{fig:sensD}. Note that unlike ethnicity, insurance type was not included in our feature sets. Self-pay patients and government insurance patients have highly erratic mortality results due to the lack of training samples. Models deteriorated slower for individuals with private insurance as opposed to individuals with Medicaid. The deterioration of performance (both gradual, and over the CareVue/MetaVision shift) is shown in Figure ~\ref{fig:sensD}.

\end{document}